\pdfoutput=1

\documentclass[11pt]{article}

\usepackage[preprint]{acl}

\usepackage{times}
\usepackage{latexsym}

\usepackage[T1]{fontenc}

\usepackage[utf8]{inputenc}

\usepackage{microtype}

\usepackage{inconsolata}

\usepackage{graphicx}

\usepackage{xspace}

\usepackage{amsmath}
\DeclareMathOperator*{\argmax}{arg\,max}
\DeclareMathOperator*{\argmin}{arg\,min}

\usepackage{listings}
\usepackage{booktabs}
\usepackage{caption}
\usepackage{subcaption}
\usepackage{amsfonts}
\usepackage{fancyvrb}
\VerbatimFootnotes 

\title{Aligning Black-box Language Models with Human Judgments}

\author{
    Gerrit J.~J. van den Burg, Gen Suzuki, Wei Liu, Murat Sensoy \\
    \{\texttt{gvdburg},\ \texttt{gfsuzuki},\ \texttt{weliuz},\ \texttt{msensoy}\}\texttt{@amazon.co.uk} \\
    Amazon
}

\begin{document}
\maketitle

\begin{abstract} 
Large language models (LLMs) are increasingly used as automated judges to evaluate recommendation systems, search engines, and other subjective tasks, where relying on human evaluators can be costly, time-consuming, and unscalable. LLMs offer an efficient solution for continuous, automated evaluation. However, since the systems that are built and improved with these judgments are ultimately designed for human use, it is crucial that LLM judgments align closely with human evaluators to ensure such systems remain human-centered.
On the other hand, aligning LLM judgments with human evaluators is challenging due to individual variability and biases in human judgments.
We propose a simple yet effective framework to align LLM judgments with individual human evaluators or their aggregated judgments, without retraining or fine-tuning the LLM. Our approach learns a linear mapping between the LLM's outputs and human judgments, achieving over $142$\% average improvement in agreement across $29$ tasks with only a small number of calibration examples used for training. 
Notably, our method works in zero-shot and few-shot settings, exceeds inter-human agreement on four out of six tasks, and enables smaller LLMs to achieve performance comparable to that of larger models.
\end{abstract}

\section{Introduction}
\label{sec:introduction}

Recent improvements to the reasoning capabilities of large language models (LLMs) have increased their use for judgment and evaluation tasks that would previously have been addressed with human evaluators~\cite{zheng2023judging,chiang2023large}. For example, LLMs have been used for judging the relevance in information retrieval systems \citep{faggioli2023perspectives}, the coherence of written discourse \citep{naismith2023automated}, or the quality of translations \citep{kocmi2023large}, among many others. A common theme in these settings is that LLMs are performing a grading task where the output label is on an ordinal scale. However, recent work has shown that the performance of LLMs on such judgment tasks varies widely depending on the specific task and the specific LLM used \citep{bavaresco2024llms}.

\begin{figure*}
    \centering
    \begin{subfigure}[b]{0.45\textwidth}
        \centering
        \includegraphics[width=\textwidth]{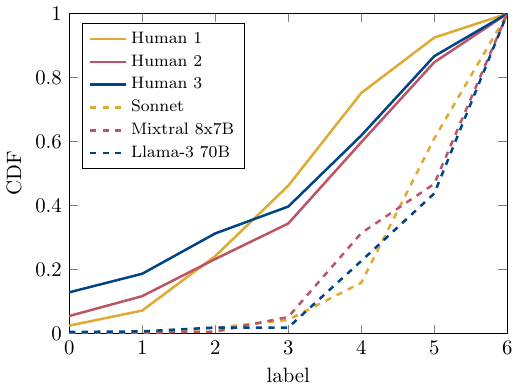}
        \caption{Non-aligned}
        \label{fig:non-aligned}
    \end{subfigure}
    \hfill
    \begin{subfigure}[b]{0.45\textwidth}
        \centering
        \includegraphics[width=\textwidth]{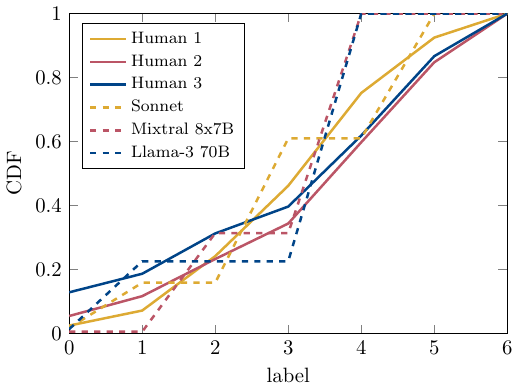}
        \caption{Aligned}
        \label{fig:aligned}
    \end{subfigure}
    \caption{Cumulative distribution function of response options for the judgment task of \citet{freitag2021experts}, before and after label alignment. The task is to grade translation quality on a rating scale from 0 to 6, i.e., \textit{nonsense} ($0$), \textit{some meaning preserved} ($2$), \textit{most meaning preserved} ($4$), and \textit{perfect} ($6$). Figure (a) illustrates the different response styles by human and LLM judges and highlights that LLMs primarily use highly positive labels, in contrast to human evaluators. Figure (b) shows the same graph after aligning the LLM responses to the average human judgment using our approach, and clearly demonstrates that we can align LLM judgments to human ones. \label{fig:cdf}}
\end{figure*}

We argue that the performance of LLMs on such judgment tasks does not automatically align well with that of human evaluators. Moreover, LLMs can exhibit their own response styles on these judgment tasks, which can be overly positive, in part due to the nature of supervised fine-tuning during their training regime. Indeed, as we show in Figure~\ref{fig:non-aligned}, LLMs can avoid negative judgments entirely on tasks where humans show more diverse judgments. Thus, an additional step is needed to align the judgment labels of the LLM with that of a human evaluators. In this work, we propose a simple but effective approach to align the LLM outputs with those of human annotators using only a small set of labeled data. We show on $29$ tasks that this additional alignment step significantly improves the performance of LLM judgments on the test set.
Figure~\ref{fig:aligned} clearly indicates that the judgment distributions of LLMs becomes much closer to that of the human evaluators after our alignment approach. 

Specifically, we propose to learn a mapping of the LLM outputs to human judgments to better align with those on a training set. In contrast to previous work on LLM calibration~\cite{zhao2021calibrate,han2023prototypical,fei-etal-2023-mitigating,zhou2024batch}, our technique is applicable without access to the model logits, which makes it much easier to use with general-purpose black-box LLMs available to the public. Moreover, we show that with our alignment approach we can bring the performance of smaller LLMs on par with that of larger ones, which can significantly reduce the cost of using LLM judges at scale.

Our key contribution is a simple yet effective approach to learn a linear mapping between a language model's categorical outputs and human judgments, enabling the alignment of the language model's judgments with individual human evaluators or aggregated majority votes. Our main contributions can be listed as follows:
\begin{itemize} 
\item We demonstrate that zero-shot and few-shot LLM judgments, without alignment, exhibit significant discrepancies compared to human evaluations, highlighting the need for additional calibration.
\item We propose a framework to align black-box LLM judgments with human evaluations on subjective tasks using a linear mapping, requiring no tuning or retraining of the LLM, and no access to model logits.
\item Our method effectively aligns LLM outputs to human judgments using a small number of calibration examples, achieving a $142$\% average improvement across $29$ tasks. Furthermore, our approach benefits smaller LLMs, enabling their use for evaluation tasks with performance comparable to larger models. 
\item The alignment outperforms inter-human consistency on four out of six tasks that involve multiple human evaluators.
\end{itemize}

The remainder of this paper is organised as follows. In Section~\ref{sec:methodology} we introduce our black-box approach for LLM alignment, and in Section~\ref{sec:experiments} we evaluate this approach through extensive and detailed experiments on a large collection of datasets. Next, in Section~\ref{sec:literature_review} we review related work. Finally, in Sections~\ref{sec:discussion} and \ref{sec:limitation} we discuss our contributions and highlight the limitations of our approach along with suggestions for future work.

\section{Methodology}
\label{sec:methodology}
Our approach aims to align the judgments provided by an LLM with those given by human evaluators, even when their respective output spaces differ. This alignment process is done without modifying the underlying LLM, making it efficient and easy to apply in practice.

\subsection{Problem Formulation}

Consider a judgment task where, for a given input instance $x_i$, the goal is to predict an output label $y_i \in \mathcal{Y}$, where $\mathcal{Y}$ is the set of possible labels given by human evaluators. For instance, in a book review evaluation task, human judgments may take on values from $\mathcal{Y} = \{\textit{bad}, \textit{good}, \textit{average}\}$, implying $|\mathcal{Y}| = 3$. We assume we have a small training dataset $\{(x_i, y_i)\}_{i=1}^N$ containing instances $x_i$ and their corresponding human judgments $y_i$.

On each task instance $x_i$ we prompt an LLM to provide its own judgment, potentially with a more nuanced set of options, denoted by $\mathcal{Z}$. For example, the LLM might use labels $\mathcal{Z} = \{\textit{bad}, \textit{neutral}, \textit{good}, \textit{excellent}\}$ implying $|\mathcal{Z}| = 4$ or stars between 1 and 5 implying $|\mathcal{Z}| = 5$. Thus, the output spaces $\mathcal{Y}$ and $\mathcal{Z}$ may differ both in size and interpretation, which allows additional flexibility in how the LLM is used while keeping the same human evaluations.

\subsection{One-hot Encoding Representation}

To facilitate alignment, we convert both human and LLM judgments into one-hot encoded vectors. Specifically, for a judgment $y_i \in \mathcal{Y}$ by a human evaluator, we represent it as a one-hot vector $\mathbf{y}_i \in \{0, 1\}^{n}$, where $n = |\mathcal{Y}|$. Similarly, for a judgment $z_i \in \mathcal{Z}$ by the LLM, we have $\mathbf{z}_i \in \{0, 1\}^{m}$, with $m = |\mathcal{Z}|$. 
For a dataset of $N$ instances, we denote the matrices of one-hot representations as:
\begin{align*}
\mathbf{Y} &= [\mathbf{y}_1, \mathbf{y}_2, \dots, \mathbf{y}_N]^\top \in \{0, 1\}^{N \times n}\\
\mathbf{Z} &= [\mathbf{z}_1, \mathbf{z}_2, \dots, \mathbf{z}_N]^\top \in \{0, 1\}^{N \times m}.
\end{align*}

\subsection{Learning the Mapping}
Our goal is to learn a mapping $\varphi : \mathcal{Z} \rightarrow \mathcal{Y}$, from the LLM's output space $\mathcal{Z}$ to the human judgment space $\mathcal{Y}$, such that the transformed LLM judgments align closely with human labels. To achieve this, we define a linear transformation matrix $\mathbf{W} \in \mathbb{R}^{m \times n}$, which maps the one-hot encoded LLM judgments to the human ones. The optimal matrix $\widehat{\mathbf{W}}$ is obtained by solving the following regularized least squares problem:
\begin{equation}
    \widehat{\mathbf{W}} = \argmin_{\mathbf{W}} \left\| \mathbf{Z} \mathbf{W} - \mathbf{Y} \right\|^2_F + \lambda \left\| \mathbf{W} \right\|^2_F,
\end{equation}
where $\left\|\cdot\right\|_F$ denotes the Frobenius norm, and $\lambda > 0$ is a regularization parameter to prevent overfitting and to avoid a singular system, as in ridge regression \citep{hoerl1970ridge}. The closed-form solution to this optimization problem is given by:
\begin{equation}
    \label{eq:w-hat}
    \widehat{\mathbf{W}} = (\mathbf{Z}^\top \mathbf{Z} + \lambda \mathbf{I})^{-1} \mathbf{Z}^\top \mathbf{Y},    
\end{equation}
where $\mathbf{I} \in \{0, 1\}^{m \times m}$ is the identity matrix.

\subsection{Aligned Judgment Inference}
Once we have the learned transformation matrix $\widehat{\mathbf{W}}$, we can align one-hot representation of an LLM's judgment $\mathbf{z}$ for a new instance by:
\begin{equation}
\varphi(\mathbf{z}) = \argmax_{j=1,\ldots, n} \{ \mathbf{z}^\top \widehat{\mathbf{W}} \}_j,
\end{equation}
where $\left\{ \mathbf{z}^\top \widehat{\mathbf{W}} \right\}_j$ represents the transformed value for each human judgment. The aligned LLM judgment is determined by selecting the human judgment corresponding to the highest transformed value, reflecting the human judgment that the alignment process associates most closely with the LLM's original output.

This approach allows us to flexibly map LLM judgments to human judgments, regardless of differences in output space sizes or interpretations. For example, if the LLM output space consists of $4$ labels while the human output space has only $3$, the learned transformation matrix $\mathbf{W} \in \mathbb{R}^{4 \times 3}$ informs a projection that aligns LLM outputs with human annotations.
This flexibility remains crucial even when the output spaces of LLMs and humans are identical, as the interpretation of the same labels can differ significantly, particularly in highly subjective judgment tasks.

\begin{table*}[t]
\centering
\footnotesize
\begin{tabular}{@{}lcrrrrrr@{}}
\toprule
 &  & \multicolumn{3}{c}{\textbf{Non-aligned}} & \multicolumn{3}{c}{\textbf{Aligned}} \\
\cmidrule(lr){3-5} \cmidrule(lr){6-8}
 &  & {\scshape\scriptsize Claude-3} & {\scshape\scriptsize Mixtral 8x7B} & {\scshape\scriptsize Llama-3 70B} & {\scshape\scriptsize Claude-3} & {\scshape\scriptsize Mixtral 8x7B} & {\scshape\scriptsize Llama-3 70B} \\
\textbf{Task Name} & {{\scshape\scriptsize Human}} & {\scshape\scriptsize Sonnet} & {\scshape\scriptsize Instruct} & {\scshape\scriptsize Instruct} & {\scshape\scriptsize Sonnet} & {\scshape\scriptsize Instruct} & {\scshape\scriptsize Instruct} \\
\midrule
Feedback-QA & 44.50 & 42.11 {\tiny $\pm 2.16$} & 43.06 {\tiny $\pm 1.57$} & 47.38 {\tiny $\pm 2.10$} & \underline{\textbf{52.68}} {\tiny $\pm 1.97$} & \underline{52.29} {\tiny $\pm 2.12$} & \underline{51.67} {\tiny $\pm 1.99$} \\
LLMBar Natural & -- & \underline{85.07} {\tiny $\pm 1.67$} & \underline{77.20} {\tiny $\pm 2.19$} & \underline{\textbf{86.52}} {\tiny $\pm 1.21$} & \underline{85.07} {\tiny $\pm 1.67$} & \underline{77.20} {\tiny $\pm 2.19$} & \underline{\textbf{86.52}} {\tiny $\pm 1.21$} \\
Medical Safety \\
\quad \emph{Query Risk Level} & -- & 39.67 {\tiny $\pm 1.15$} & 43.20 {\tiny $\pm 1.30$} & 21.12 {\tiny $\pm 1.80$} & \underline{85.00} {\tiny $\pm 1.47$} & \underline{\textbf{85.30}} {\tiny $\pm 1.57$} & \underline{85.22} {\tiny $\pm 1.45$} \\
\quad \emph{Response Type} & -- & 5.63 {\tiny $\pm 0.52$} & 11.03 {\tiny $\pm 1.31$} & 5.63 {\tiny $\pm 1.00$} & \underline{\textbf{79.92}} {\tiny $\pm 1.87$} & \underline{68.14} {\tiny $\pm 3.23$} & \underline{75.08} {\tiny $\pm 1.80$} \\
Newsroom \\
\quad \emph{Coherence} & 24.29 & 27.87 {\tiny $\pm 0.96$} & 26.11 {\tiny $\pm 0.50$} & 31.94 {\tiny $\pm 0.58$} & \underline{32.24} {\tiny $\pm 1.15$} & \underline{30.60} {\tiny $\pm 1.98$} & \underline{\textbf{32.73}} {\tiny $\pm 2.12$} \\
\quad \emph{Fluency} & 21.35 & \underline{28.03} {\tiny $\pm 0.65$} & 20.27 {\tiny $\pm 0.82$} & \underline{\textbf{30.70}} {\tiny $\pm 0.70$} & 25.88 {\tiny $\pm 1.44$} & \underline{24.01} {\tiny $\pm 2.07$} & 29.83 {\tiny $\pm 0.87$} \\
\quad \emph{Informativeness} & 31.75 & 32.77 {\tiny $\pm 0.86$} & 15.53 {\tiny $\pm 1.21$} & 35.01 {\tiny $\pm 0.91$} & \underline{38.32} {\tiny $\pm 1.50$} & \underline{29.94} {\tiny $\pm 3.18$} & \underline{\textbf{38.97}} {\tiny $\pm 2.14$} \\
\quad \emph{Relevance} & 30.71 & 30.96 {\tiny $\pm 0.93$} & 21.24 {\tiny $\pm 0.58$} & 33.58 {\tiny $\pm 0.68$} & \underline{36.04} {\tiny $\pm 1.79$} & \underline{35.57} {\tiny $\pm 1.68$} & \underline{\textbf{38.42}} {\tiny $\pm 2.34$} \\
ROSCOE-CosmosQA \\
\quad \emph{Coherency} & -- & 42.65 {\tiny $\pm 1.90$} & 20.41 {\tiny $\pm 1.61$} & 36.92 {\tiny $\pm 1.95$} & \underline{47.96} {\tiny $\pm 3.69$} & \underline{42.99} {\tiny $\pm 5.57$} & \underline{\textbf{53.17}} {\tiny $\pm 4.43$} \\
\quad \emph{Contradiction} & -- & 68.64 {\tiny $\pm 2.54$} & 57.20 {\tiny $\pm 1.22$} & 61.55 {\tiny $\pm 2.17$} & \underline{\textbf{78.03}} {\tiny $\pm 0.96$} & \underline{77.94} {\tiny $\pm 0.98$} & \underline{77.97} {\tiny $\pm 0.99$} \\
\quad \emph{Missing Steps} & -- & \underline{59.12} {\tiny $\pm 1.76$} & \underline{56.73} {\tiny $\pm 1.37$} & \underline{\textbf{61.72}} {\tiny $\pm 2.20$} & 55.99 {\tiny $\pm 2.56$} & 55.31 {\tiny $\pm 1.72$} & 60.81 {\tiny $\pm 2.50$} \\
\quad \emph{Overall Quality} & -- & 29.86 {\tiny $\pm 2.38$} & 18.50 {\tiny $\pm 1.48$} & 33.47 {\tiny $\pm 1.98$} & \underline{43.47} {\tiny $\pm 2.48$} & \underline{31.90} {\tiny $\pm 2.84$} & \underline{\textbf{44.66}} {\tiny $\pm 3.60$} \\
ROSCOE-DROP \\
\quad \emph{Coherency} & -- & 72.66 {\tiny $\pm 1.98$} & 34.68 {\tiny $\pm 2.40$} & 67.56 {\tiny $\pm 1.75$} & \underline{\textbf{77.59}} {\tiny $\pm 2.20$} & \underline{77.22} {\tiny $\pm 1.60$} & \underline{75.67} {\tiny $\pm 1.71$} \\
\quad \emph{Contradiction} & -- & 88.16 {\tiny $\pm 0.70$} & 71.24 {\tiny $\pm 1.39$} & 86.48 {\tiny $\pm 0.96$} & \underline{93.35} {\tiny $\pm 1.84$} & \underline{94.03} {\tiny $\pm 0.65$} & \underline{\textbf{94.54}} {\tiny $\pm 0.74$} \\
\quad \emph{Missing Steps} & -- & \underline{51.46} {\tiny $\pm 1.70$} & \underline{51.58} {\tiny $\pm 1.34$} & \underline{\textbf{55.44}} {\tiny $\pm 1.59$} & 50.38 {\tiny $\pm 1.07$} & 48.68 {\tiny $\pm 2.26$} & 51.49 {\tiny $\pm 3.53$} \\
\quad \emph{Overall Quality} & -- & 43.42 {\tiny $\pm 2.14$} & \underline{40.89} {\tiny $\pm 1.82$} & 43.53 {\tiny $\pm 2.20$} & \underline{43.67} {\tiny $\pm 8.60$} & 39.62 {\tiny $\pm 2.06$} & \underline{\textbf{46.75}} {\tiny $\pm 3.26$} \\
ROSCOE-ESNLI \\
\quad \emph{Coherency} & -- & 84.65 {\tiny $\pm 1.19$} & 39.39 {\tiny $\pm 2.06$} & 73.59 {\tiny $\pm 1.80$} & \underline{86.49} {\tiny $\pm 1.58$} & \underline{89.04} {\tiny $\pm 0.90$} & \underline{\textbf{89.31}} {\tiny $\pm 2.27$} \\
\quad \emph{Contradiction} & -- & 93.25 {\tiny $\pm 0.96$} & 50.57 {\tiny $\pm 2.37$} & 93.46 {\tiny $\pm 0.99$} & \underline{95.26} {\tiny $\pm 2.19$} & \underline{\textbf{96.55}} {\tiny $\pm 0.84$} & \underline{94.16} {\tiny $\pm 1.78$} \\
\quad \emph{Missing Steps} & -- & 67.98 {\tiny $\pm 1.97$} & 27.28 {\tiny $\pm 1.38$} & 64.26 {\tiny $\pm 1.47$} & \underline{\textbf{74.21}} {\tiny $\pm 1.72$} & \underline{73.16} {\tiny $\pm 1.43$} & \underline{72.80} {\tiny $\pm 3.23$} \\
\quad \emph{Overall Quality} & -- & \underline{\textbf{48.07}} {\tiny $\pm 2.63$} & \underline{38.86} {\tiny $\pm 2.26$} & \underline{47.55} {\tiny $\pm 2.73$} & 45.00 {\tiny $\pm 6.35$} & 36.49 {\tiny $\pm 6.93$} & 43.69 {\tiny $\pm 6.75$} \\
ROSCOE-GSM8K \\
\quad \emph{Coherency} & -- & 59.73 {\tiny $\pm 1.84$} & 47.50 {\tiny $\pm 2.70$} & \underline{\textbf{63.89}} {\tiny $\pm 1.83$} & \underline{62.07} {\tiny $\pm 1.31$} & \underline{60.35} {\tiny $\pm 3.32$} & 62.94 {\tiny $\pm 1.79$} \\
\quad \emph{Contradiction} & -- & 80.67 {\tiny $\pm 0.94$} & 55.30 {\tiny $\pm 2.37$} & 80.53 {\tiny $\pm 1.42$} & \underline{83.60} {\tiny $\pm 1.58$} & \underline{83.32} {\tiny $\pm 1.61$} & \underline{\textbf{83.81}} {\tiny $\pm 1.90$} \\
\quad \emph{Missing Steps} & -- & \underline{81.13} {\tiny $\pm 1.46$} & 42.77 {\tiny $\pm 0.77$} & \underline{\textbf{84.09}} {\tiny $\pm 1.26$} & \underline{81.13} {\tiny $\pm 1.46$} & \underline{58.89} {\tiny $\pm 1.97$} & \underline{\textbf{84.09}} {\tiny $\pm 1.26$} \\
\quad \emph{Overall Quality} & -- & 71.60 {\tiny $\pm 2.15$} & 49.59 {\tiny $\pm 2.37$} & 65.46 {\tiny $\pm 2.03$} & \underline{72.53} {\tiny $\pm 2.71$} & \underline{61.61} {\tiny $\pm 3.42$} & \underline{\textbf{73.82}} {\tiny $\pm 2.86$} \\
SummEval \\
\quad \emph{Coherence} & -- & 31.47 {\tiny $\pm 1.71$} & 28.79 {\tiny $\pm 2.06$} & \underline{\textbf{40.90}} {\tiny $\pm 1.48$} & \underline{40.20} {\tiny $\pm 1.69$} & \underline{31.64} {\tiny $\pm 2.45$} & 39.87 {\tiny $\pm 2.44$} \\
\quad \emph{Consistency} & -- & 83.64 {\tiny $\pm 1.53$} & 68.84 {\tiny $\pm 1.80$} & 68.82 {\tiny $\pm 2.53$} & \underline{\textbf{85.38}} {\tiny $\pm 1.55$} & \underline{84.44} {\tiny $\pm 1.80$} & \underline{77.83} {\tiny $\pm 4.43$} \\
\quad \emph{Fluency} & -- & 4.50 {\tiny $\pm 0.83$} & 3.33 {\tiny $\pm 0.58$} & 2.58 {\tiny $\pm 0.33$} & \underline{79.73} {\tiny $\pm 1.54$} & \underline{79.70} {\tiny $\pm 1.35$} & \underline{\textbf{80.22}} {\tiny $\pm 1.70$} \\
\quad \emph{Relevance} & -- & 37.27 {\tiny $\pm 1.53$} & 31.27 {\tiny $\pm 1.72$} & 41.34 {\tiny $\pm 1.40$} & \underline{\textbf{55.07}} {\tiny $\pm 4.67$} & \underline{51.30} {\tiny $\pm 3.72$} & \underline{54.17} {\tiny $\pm 3.01$} \\
WMT-20-EnDe & 30.02 & 19.48 {\tiny $\pm 1.45$} & 17.58 {\tiny $\pm 1.58$} & 17.15 {\tiny $\pm 1.45$} & \underline{\textbf{26.70}} {\tiny $\pm 1.80$} & \underline{25.05} {\tiny $\pm 1.33$} & \underline{23.76} {\tiny $\pm 1.37$} \\
\bottomrule
\end{tabular}
\caption{Summary table showing the accuracy (in \%) of aligning LLMs to human labels. The first column shows inter-human agreement for the tasks with more than one human annotator. For the non-aligned results the label predicted by the LLM is used directly. For each task, $25$\% labeled samples are used for alignment and $75$\% for evaluation, and accuracy is averaged over each individual human annotator. Experiments are repeated $10$ times with different training and evaluation splits and results are averaged. \textbf{Bold} indicates the best performance in the each row and  \underline{underscore} indicates the best performance of each LLM (aligned or non-aligned). \label{tab:alignment_results} }
\end{table*}

\section{Experiments}
\label{sec:experiments}
To evaluate the effectiveness of our proposed alignment approach, we conduct a series of experiments across $29$ tasks with three widely-adopted large language models: Claude-3 Sonnet \citep{anthropic2024claude3}, Mixtral 8x7B Instruct \citep{jiang2024mixtral}, and Llama-3 70B  Instruct \cite{dubey2024llama}. With the exception of the Feedback-QA dataset \citep{li2022using}, each of the datasets and prompts used to query LLMs are obtained from Judge-Bench, a benchmark dataset proposed recently by \citet{bavaresco2024llms} to evaluate LLMs as judges.
The datasets we consider from Judge-Bench include LLMBar \citep{zeng2024evaluating}, Medical-Safety \citep{abercrombie2022risk}, Newsroom \citep{grusky2018newsroom},  SummEval \citep{liu2023geval}, WMT-20-EnDe \citep{freitag2021experts}, and the ROSCOE metrics \citep{golovneva2023roscoe} on COSMOS-QA \citep{huang2019cosmos}, DROP \citep{dua2019drop}, ESNLI \citep{camburu2018esnli}, and GSM8K \citep{cobbe2021training}. Since many datasets have multiple metrics for evaluation, this results in a total of 29 tasks.

In our evaluations, we primarily focused on the case where both LLMs and human evaluators operate within the same judgment space, implying that $\mathcal{Z} = \mathcal{Y}$. This setup allows us to directly assess and discuss the impact of alignment on the agreement between LLM and human judgments. Notably, if $\mathcal{Z} \neq \mathcal{Y}$, the agreement metric would not be straightforward to calculate without an alignment step.
In our experiments, we set $\lambda = 10^{-6}$ in eq.~(\ref{eq:w-hat}), as it is primarily used to avoid issues with matrix inversion.
    
Our primary objective was to assess the agreement between human judgments and those generated by the LLMs. We treated human judgments as the ground truth and calculated the accuracy of the LLM outputs relative to these targets. Additionally, for task datasets with multiple human evaluators, we analyzed the inter-human agreement. In our experimental setup, for each task, we used randomly selected $100$ examples for training the alignment model and $300$ examples for testing the effectiveness of the learned alignment. Because the labels differ between tasks, alignment mappings are learned on a per-task basis (we study transferring alignments between tasks in Section~\ref{sec:transfer}).
In some tasks, the total number of examples are below $400$; so we use $25$\% of the task dataset as training and $75$\% for testing. 
We repeated our experiments $10$ times with different random splits of training and test sets, and report the mean and standard deviation.

\subsection{Zero-shot Judgments}
In Table \ref{tab:alignment_results}, we present the accuracy of LLM judgments compared to human ones, both before and after the alignment process. The results indicate a significant increase in agreement post-alignment, demonstrating the effectiveness of our approach in enhancing LLM performance in alignment tasks.

Overall, our approach significantly improves LLM-to-human agreement or maintains it at the same level. There are only a few instances where alignment results in a slight reduction in agreement. On average, across all $29$ tasks, alignment leads to a remarkable 142\% increase in agreement. More specifically, we have achieved $116.52$\% increase for Claude-3 Sonnet, $142.91$\% for Mixtral 8x7B, and $166.6$\% for Llama-3 70B models. 
Note that we observe identical results before and after alignment for some tasks or models, because our alignment results in the identity mapping in these cases.

For specific tasks, such as \textit{Medical Safety (response type)} and \textit{Summeval (fluency)}, the agreement increases from single-digit percentages up to $80$\% after alignment.
For example, we observed a significant discrepancy between the LLM judgments and evaluations made by a professional nurse in the medical safety task. While the nurse often selected \textit{``not medical''} as the judgment for many cases, the LLMs tended to choose \textit{``non-serious''}. Such misalignment not only led to lower agreement scores but also highlighted the potential risks of employing LLMs as judges in critical domains, such as medical assessments.
Our alignment method effectively learned to map the LLM's \textit{``non-serious''} outputs to \textit{``not medical''} judgments of the nurses. As a result of these learned mappings, we improved the agreement between LLMs and nurses from 5-11\% up to 80\% for this task.

\begin{figure}
    \centering
    \includegraphics[width=1.\linewidth]{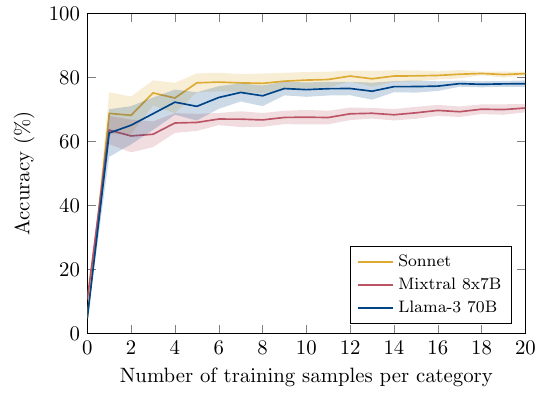}
    \caption{Test accuracy for \textit{Medical Safety (response type)} dataset as we increase the number of training examples per judgment category.}
    \label{fig:medical_safety}
\end{figure}

In our experiments, we used up to $100$ training examples to learn the alignment. However, we observed that our approach works quite well with much smaller numbers of training examples across all datasets.
Figure~\ref{fig:medical_safety} shows the test accuracy of our approach for the \textit{Medical Safety (response type)} task for different numbers of training examples per judgment category. 
The figure indicates that our approach achieves alignment and significantly improves the agreement between the LLMs and human judgments using only one or two examples per category.
This highlights the sample efficiency of our approach, which is in part due to its simplicity.

\begin{table*}[t]
\centering
\footnotesize
\begin{tabular}{@{}lrrrrrr@{}}
\toprule
 & \multicolumn{3}{c}{\textbf{Non-aligned}} & \multicolumn{3}{c}{\textbf{Aligned}} \\
\cmidrule(lr){2-4} \cmidrule(lr){5-7}
 & {\scshape\scriptsize Claude-3} & {\scshape\scriptsize Mixtral 8x7B} & {\scshape\scriptsize Llama-3 70B} & {\scshape\scriptsize Claude-3} & {\scshape\scriptsize Mixtral 8x7B} & {\scshape\scriptsize Llama-3 70B} \\
\textbf{Task Name} & {\scshape\scriptsize Sonnet} & {\scshape\scriptsize Instruct} & {\scshape\scriptsize Instruct} & {\scshape\scriptsize Sonnet} & {\scshape\scriptsize Instruct} & {\scshape\scriptsize Instruct} \\
\midrule
Feedback-QA & 43.29 {\tiny $\pm 1.01$} & 40.28 {\tiny $\pm 0.82$} & 45.17 {\tiny $\pm 1.61$} & \underline{52.63} {\tiny $\pm 0.91$} & \underline{50.63} {\tiny $\pm 1.35$} & \underline{\textbf{53.43}} {\tiny $\pm 1.34$} \\
LLMBar Natural & \underline{\textbf{84.67}} {\tiny $\pm 2.09$} & \underline{71.83} {\tiny $\pm 2.86$} & \underline{81.11} {\tiny $\pm 1.89$} & \underline{\textbf{84.67}} {\tiny $\pm 2.09$} & \underline{71.83} {\tiny $\pm 2.86$} & \underline{81.11} {\tiny $\pm 1.89$} \\
Medical Safety \\
\quad \emph{Query Risk Level} & 46.73 {\tiny $\pm 1.53$} & 32.32 {\tiny $\pm 1.46$} & 23.90 {\tiny $\pm 1.03$} & \underline{85.73} {\tiny $\pm 1.03$} & \underline{85.21} {\tiny $\pm 2.73$} & \underline{\textbf{86.01}} {\tiny $\pm 1.32$} \\
\quad \emph{Response Type} & 8.62 {\tiny $\pm 0.84$} & 11.61 {\tiny $\pm 1.07$} & 10.20 {\tiny $\pm 0.53$} & \underline{\textbf{76.80}} {\tiny $\pm 2.48$} & \underline{69.97} {\tiny $\pm 2.16$} & \underline{76.49} {\tiny $\pm 2.56$} \\
Newsroom \\
\quad \emph{Coherence} & 25.63 {\tiny $\pm 1.05$} & 18.47 {\tiny $\pm 0.82$} & 25.04 {\tiny $\pm 0.73$} & \underline{30.22} {\tiny $\pm 0.94$} & \underline{\textbf{31.12}} {\tiny $\pm 1.05$} & \underline{30.07} {\tiny $\pm 1.08$} \\
\quad \emph{Fluency} & \underline{26.60} {\tiny $\pm 0.55$} & 20.21 {\tiny $\pm 0.91$} & \underline{\textbf{29.62}} {\tiny $\pm 0.56$} & 24.88 {\tiny $\pm 1.50$} & \underline{24.32} {\tiny $\pm 1.71$} & 27.35 {\tiny $\pm 1.28$} \\
\quad \emph{Informativeness} & 28.70 {\tiny $\pm 0.70$} & 18.94 {\tiny $\pm 1.27$} & 27.87 {\tiny $\pm 1.07$} & \underline{37.68} {\tiny $\pm 1.69$} & \underline{31.48} {\tiny $\pm 1.71$} & \underline{\textbf{37.84}} {\tiny $\pm 1.80$} \\
\quad \emph{Relevance} & 30.64 {\tiny $\pm 1.28$} & 14.25 {\tiny $\pm 0.58$} & 32.95 {\tiny $\pm 0.85$} & \underline{35.27} {\tiny $\pm 2.74$} & \underline{35.71} {\tiny $\pm 1.03$} & \underline{\textbf{36.52}} {\tiny $\pm 1.14$} \\
ROSCOE-CosmosQA \\
\quad \emph{Coherency} & 43.91 {\tiny $\pm 1.53$} & 53.74 {\tiny $\pm 1.10$} & 42.19 {\tiny $\pm 2.46$} & \underline{53.49} {\tiny $\pm 2.26$} & \underline{\textbf{54.49}} {\tiny $\pm 2.78$} & \underline{46.26} {\tiny $\pm 1.80$} \\
\quad \emph{Contradiction} & 77.28 {\tiny $\pm 1.78$} & 47.76 {\tiny $\pm 2.08$} & 57.90 {\tiny $\pm 1.72$} & \underline{77.41} {\tiny $\pm 1.72$} & \underline{\textbf{78.50}} {\tiny $\pm 1.47$} & \underline{78.09} {\tiny $\pm 1.46$} \\
\quad \emph{Missing Steps} & \underline{55.44} {\tiny $\pm 1.70$} & \underline{56.94} {\tiny $\pm 1.36$} & \underline{\textbf{67.62}} {\tiny $\pm 1.62$} & 54.49 {\tiny $\pm 1.27$} & 55.71 {\tiny $\pm 1.56$} & \underline{\textbf{67.62}} {\tiny $\pm 1.62$} \\
\quad \emph{Overall Quality} & 31.54 {\tiny $\pm 2.80$} & 33.74 {\tiny $\pm 2.16$} & 42.88 {\tiny $\pm 2.38$} & \underline{44.60} {\tiny $\pm 2.14$} & \underline{45.44} {\tiny $\pm 5.12$} & \underline{\textbf{45.80}} {\tiny $\pm 3.74$} \\
ROSCOE-DROP \\
\quad \emph{Coherency} & 71.33 {\tiny $\pm 1.47$} & 74.49 {\tiny $\pm 1.60$} & 63.49 {\tiny $\pm 1.90$} & \underline{74.43} {\tiny $\pm 2.18$} & \underline{\textbf{76.33}} {\tiny $\pm 1.33$} & \underline{73.91} {\tiny $\pm 3.80$} \\
\quad \emph{Contradiction} & 91.39 {\tiny $\pm 0.95$} & 43.42 {\tiny $\pm 1.63$} & 75.63 {\tiny $\pm 1.72$} & \underline{92.59} {\tiny $\pm 1.75$} & \underline{\textbf{93.92}} {\tiny $\pm 0.42$} & \underline{93.81} {\tiny $\pm 0.43$} \\
\quad \emph{Missing Steps} & \underline{53.86} {\tiny $\pm 1.73$} & \underline{54.18} {\tiny $\pm 1.88$} & \underline{\textbf{61.77}} {\tiny $\pm 1.30$} & 51.14 {\tiny $\pm 1.67$} & 48.10 {\tiny $\pm 2.77$} & 60.24 {\tiny $\pm 4.37$} \\
\quad \emph{Overall Quality} & 45.76 {\tiny $\pm 1.44$} & 42.34 {\tiny $\pm 1.40$} & 44.88 {\tiny $\pm 1.68$} & \underline{\textbf{47.53}} {\tiny $\pm 2.32$} & \underline{43.04} {\tiny $\pm 2.10$} & \underline{45.84} {\tiny $\pm 2.20$} \\
ROSCOE-ESNLI \\
\quad \emph{Coherency} & 83.68 {\tiny $\pm 1.67$} & 86.32 {\tiny $\pm 1.05$} & 77.81 {\tiny $\pm 1.11$} & \underline{\textbf{88.42}} {\tiny $\pm 2.51$} & \underline{87.28} {\tiny $\pm 1.43$} & \underline{88.07} {\tiny $\pm 3.31$} \\
\quad \emph{Contradiction} & 95.44 {\tiny $\pm 1.23$} & 55.26 {\tiny $\pm 2.08$} & 83.51 {\tiny $\pm 1.51$} & \underline{96.67} {\tiny $\pm 1.46$} & \underline{\textbf{96.93}} {\tiny $\pm 1.13$} & \underline{\textbf{96.93}} {\tiny $\pm 1.13$} \\
\quad \emph{Missing Steps} & 70.79 {\tiny $\pm 1.84$} & 27.72 {\tiny $\pm 2.12$} & 65.07 {\tiny $\pm 2.40$} & \underline{71.05} {\tiny $\pm 1.80$} & \underline{\textbf{72.28}} {\tiny $\pm 2.12$} & \underline{70.09} {\tiny $\pm 4.37$} \\
\quad \emph{Overall Quality} & 48.07 {\tiny $\pm 1.87$} & 45.00 {\tiny $\pm 1.67$} & \underline{45.66} {\tiny $\pm 1.64$} & \underline{\textbf{48.51}} {\tiny $\pm 2.36$} & \underline{45.35} {\tiny $\pm 2.32$} & 45.40 {\tiny $\pm 2.38$} \\
ROSCOE-GSM8K \\
\quad \emph{Coherency} & 63.60 {\tiny $\pm 1.55$} & 59.07 {\tiny $\pm 2.24$} & 59.80 {\tiny $\pm 2.11$} & \underline{\textbf{66.93}} {\tiny $\pm 2.13$} & \underline{61.20} {\tiny $\pm 2.23$} & \underline{63.87} {\tiny $\pm 2.75$} \\
\quad \emph{Contradiction} & \underline{\textbf{84.53}} {\tiny $\pm 1.81$} & 45.13 {\tiny $\pm 2.77$} & 77.09 {\tiny $\pm 1.73$} & \underline{\textbf{84.53}} {\tiny $\pm 1.81$} & \underline{\textbf{84.53}} {\tiny $\pm 1.81$} & \underline{81.18} {\tiny $\pm 3.83$} \\
\quad \emph{Missing Steps} & \underline{74.87} {\tiny $\pm 1.52$} & 54.67 {\tiny $\pm 2.13$} & \underline{\textbf{94.31}} {\tiny $\pm 0.76$} & \underline{74.87} {\tiny $\pm 1.52$} & \underline{57.47} {\tiny $\pm 2.32$} & \underline{\textbf{94.31}} {\tiny $\pm 0.76$} \\
\quad \emph{Overall Quality} & 72.00 {\tiny $\pm 1.23$} & 71.47 {\tiny $\pm 1.29$} & \underline{74.31} {\tiny $\pm 1.16$} & \underline{\textbf{75.20}} {\tiny $\pm 3.66$} & \underline{71.67} {\tiny $\pm 1.67$} & 72.43 {\tiny $\pm 2.78$} \\
SummEval \\
\quad \emph{Coherence} & 25.03 {\tiny $\pm 1.41$} & 24.87 {\tiny $\pm 2.01$} & 27.33 {\tiny $\pm 1.61$} & \underline{\textbf{37.90}} {\tiny $\pm 2.24$} & \underline{31.93} {\tiny $\pm 2.92$} & \underline{34.40} {\tiny $\pm 2.29$} \\
\quad \emph{Consistency} & 75.03 {\tiny $\pm 1.38$} & 12.00 {\tiny $\pm 1.20$} & 36.61 {\tiny $\pm 1.22$} & \underline{\textbf{87.13}} {\tiny $\pm 1.72$} & \underline{83.06} {\tiny $\pm 5.03$} & \underline{86.71} {\tiny $\pm 1.56$} \\
\quad \emph{Fluency} & 35.57 {\tiny $\pm 0.78$} & 8.93 {\tiny $\pm 0.79$} & 4.55 {\tiny $\pm 0.75$} & \underline{82.47} {\tiny $\pm 1.14$} & \underline{81.30} {\tiny $\pm 3.34$} & \underline{\textbf{83.30}} {\tiny $\pm 1.59$} \\
\quad \emph{Relevance} & 25.57 {\tiny $\pm 1.13$} & 1.87 {\tiny $\pm 0.85$} & 37.79 {\tiny $\pm 1.55$} & \underline{\textbf{55.33}} {\tiny $\pm 0.95$} & \underline{42.76} {\tiny $\pm 2.63$} & \underline{52.48} {\tiny $\pm 3.21$} \\
WMT-20-EnDe & 15.25 {\tiny $\pm 0.84$} & 16.09 {\tiny $\pm 0.82$} & 24.76 {\tiny $\pm 0.68$} & \underline{\textbf{28.42}} {\tiny $\pm 1.75$} & \underline{27.33} {\tiny $\pm 1.22$} & \underline{25.20} {\tiny $\pm 1.17$} \\
\bottomrule
\end{tabular}
\caption{In-Context Learning experiments, where one example per judgment category is given to LLMs as in-context examples. Experiments are repeated 10 times with different training and evaluation splits and results are averaged. \textbf{Bold} indicates the best performance in the each row and  \underline{underscore} indicates the best performance of each LLM (aligned or non-aligned). \label{tab:in_context_results}}
\end{table*}

\subsection{Judgments with In-Context Learning}
Our findings in the zero-shot setting motivated us to investigate whether the sources of misalignment between LLMs and humans could be addressed through in-context learning \citep{brown2020language}, by providing examples of human judgments for each output label. Specifically, for each human judgment label we include in the LLM prompt an example on which human evaluators agreed, aiming to guide the LLM towards better alignment with human interpretations.
 
As shown in Table~\ref{tab:in_context_results}, the results on this experiment indicate that merely providing these examples did not lead to significantly improved agreement between the LLMs and human evaluators. 
That is, the in-context learning barely improved the agreement between human judgments and LLMs, except a few cases such as \textit{Summeval (fluency)}, where the agreement with human evaluators improved from $4.5$\% to $35.57$\% for \textit{Claude-3 Sonnet}.
The average agreement improved only $0.8$\% for Claude-3 Sonnet and $1.14$\% for Mixtral 8x7B with very high standard deviations ($7.19$\% and $19.74$\%, respectively).
We observe $1.4$\% decrease in the average performance of Llama-3 70B with $8.45$\% standard deviation.
One potential reason for poorer performance can be the increased complexity of the prompts with the added examples, which can become too long for the model to effectively follow the instructions or even approach the model's maximum context length.

When our alignment approach is used on top of in-context learning, we observe $111.74$\% average improvement with respect to non-aligned model agreement with human judgments. More specifically, we have achieved $49.65$\% increase for Claude-3 Sonnet, $180.13$\% for Mixtral 8x7B, and $105.45$\% for Llama-3 70B models.
That is, our alignment approach consistently results in significant improvements in LLMs' agreement with human judgments, in both zero-shot and in-context learning settings.

These findings underscore the necessity of a structured alignment process. The performance enhancements achieved through our proposed method surpass those observed with traditional in-context learning strategies. This suggests that while in-context examples can provide useful context for LLMs, in our experiments they do not substitute for a dedicated alignment mechanism to effectively bridge the gap between LLM judgments and human evaluations.

\begin{table}
    \footnotesize
    \centering
    \begin{tabular}{lrr}
    \toprule
    & Zero-shot & In-Context Learning \\
    \midrule
    Claude-3 Sonnet & $79.31$\%  & $79.31$\%  \\
    Mixtral 8x7B Instruct & $82.75$\%  & $89.65$\%  \\
    Llama-3 70B Instruct & $72.41$\%  & $75.86$\%  \\
    \bottomrule
    \end{tabular}
    \caption{Percentage of tasks in which the proposed approach improves the agreement with human judgments.}
    \label{tab:dataset_agrement}
\end{table}

In summary, our experiments highlight that aligning LLM judgments with human assessments is a complex task. Our proposed alignment methodology offers a promising solution despite its simplicity, even in scenarios where in-context learning is employed.
Table~\ref{tab:dataset_agrement} lists the percentage of tasks in which our approach improves the level of agreement with human judgments after alignment. It indicates that we were able to significantly improve agreement between LLMs and the human judments with our approach in both zero-shot and in-context learning settings. 
One of our most interesting and important findings is that our alignment approach has enabled smaller models like \textit{Mixtral 8x7B Instruct} to match the performance of much larger models such as \textit{Claude-3 Sonnet}.
This is a significant result, as it opens up new opportunities to use these smaller, more affordable and openly accessible language models for evaluation tasks, while still achieving similar performance to large, proprietary API-based models like those in the Claude family.

\subsection{Transferring alignments \label{sec:transfer}}
For judgment tasks that use grading scales with the same number of categories and the same meaning it is possible to evaluate whether an alignment learned on one task can be used on a different task. To test this, we learn an alignment on the coherency task of the Roscoe-Cosmos dataset, and test it on the coherency task of the other Roscoe datasets (each of which evaluates the coherence of GPT-3's response on reasoning tasks). The results in Table~\ref{tab:transfer} indicate that using the transferred alignment gives better performance than the non-aligned LLM judges in Table~\ref{tab:alignment_results}, and is only slightly worse than when using a task-specific alignment.

\begin{table}
    \footnotesize
    \centering
    \begin{tabular}{lrrr}
    \toprule
     & Claude-3 & Mixtral 8x7B & Llama-3 70B \\
     & Sonnet & Instruct & Instruct \\
    \midrule
    DROP & 76.62 & 65.98 & 75.29 \\
    ESNLI & 86.20 & 78.07 & 85.13 \\
    GSM8K & 65.11 & 51.40 & 65.79 \\
    \bottomrule
    \end{tabular}
    \caption{Accuracy (in \%) using an alignment learned on the coherency task of Roscoe-Cosmos and tested on the other Roscoe coherency tasks. As in Table~\ref{tab:alignment_results}, 25\% of training data was used to learn the mapping and results were averaged 10 times with different training data splits.}
    \label{tab:transfer}
\end{table}

\section{Related Work}
\label{sec:literature_review}
The use of large language models as automated judges has been gaining traction due to their potential for scalability and efficiency~\cite{zheng2023judging, bavaresco2024llms}. However, as these models are increasingly relied upon to evaluate recommendations, search results, and other subjective tasks, it is crucial that their judgments align closely with human evaluators to ensure evaluations remain human-centered and useful.

\citet{gilardi2023chatgpt} investigated the capabilities of ChatGPT compared to human annotators across tweet annotation tasks including \textit{relevance}, \textit{stance}, \textit{topics}, and \textit{frame detection}. The results show that ChatGPT's zero-shot accuracy exceeds that of crowd workers and echoes the findings on annotating political affiliation in tweets by \citet{tornberg2023chatgpt}.
The cost-effectiveness of ChatGPT underscores its potential to drastically improve the efficiency of text classification tasks. 
Despite of these promising initial results, recent work has highlighted the challenges in relying on LLMs as judges due to biases and inconsistencies in their judgments \citep{wu2023style,zheng2023large,koo2024benchmarking,hada2024large,pavlovic2024effectiveness}. \citet{wang2023large} demonstrated that the quality ranking of candidate responses can be manipulated by altering their order of appearance, allowing one model to seem superior to another based on positional biases. This highlights the susceptibility of LLM evaluations to manipulation and underscores the need for effective calibration to ensure fairness and reliability.

\citet{bavaresco2024llms} proposed Judge-Bench, a benchmark consisting of over 20 NLP datasets with human annotations, to assess the effectiveness of LLMs as judges. Their evaluations of $11$ different LLMs, including both open-weight and proprietary models, reveal significant variance in correlation to human judgments across datasets. They conclude that LLMs, despite their growing usage in evaluation settings, are not yet ready to systematically replace human judges.
In this paper, we also used the datasets provided by this benchmark to demonstrate the effectiveness of our approach in improving the alignment of the LLMs with the human judgments. 

The calibration of response styles has a long history in the psychology literature, where survey respondents can exhibit biases towards the central or extreme categories on a rating scale \citep{jackson1958content,paulhus1991measurement}. For such settings with a large number of respondents and heterogeneous survey questions, techniques have been developed for identifying and removing response style bias \citep[see, e.g.,][]{van2010identifying,schoonees2015constrained}. This echoes the observations reported in this work, where we have shown that LLMs exhibit their own response styles, for instance by preferring highly positive classes (see Figure~\ref{fig:cdf}). At the same time, \citet{huang-etal-2024-chatgpt} demonstrated that on coarse-grained assessments, such as binary assessments, ChatGPT’s evaluations
closely approximate human ones, but that it struggles in fine-grained assessments. On the other hand, \citet{tjuatja2024llms} found that LLMs do not exhibit human-like response style biases caused by the \emph{wording} of a prompt. 

When the logits of language models are accessible, they can be calibrated directly using gradient descent as in model fine-tuning or low-rank adaptation \cite{reif2024beyond}. \citet{zhao2021calibrate} showed that LLMs are highly sensitive to prompt structure, resulting in inconsistent performance, and proposed a calibration procedure involving the use of a content-free input to estimate and adjust for these biases. Such findings highlight the importance of addressing LLM biases, especially in contexts where alignment with human judgments is critical.
Unlike the above approaches, our method does not require access to model weights or logits, so it is easier to adapt to a wider range of use cases and scenarios.

\section{Discussion}
\label{sec:discussion}
Our method provides a simple yet powerful framework to align LLM judgments with human evaluations, offering substantial improvements without the need for model retraining or fine-tuning. By learning linear mappings between LLM outputs and human judgments, we achieve significant improvement in agreement, a result that is both significant and practical for real-world applications where human labeled data is scarce. Furthermore, in four out of six tasks where there are multiple human annotators, the LLM performance after our alignment exceeds the inter-human agreement.

The simplicity and scalability of our approach is a key advantage. With only minimal calibration examples, we allow smaller models to perform at levels comparable to much larger, more resource-intensive LLMs. This shows that our method may greatly extend the usefulness of smaller models in judgment tasks, opening up opportunities for their deployment in a wider range of applications. This is especially valuable in resource-constrained settings, where using large models may be impractical or too costly.

Although this extension is beyond the scope of the current paper, we note that when model logits are available, the one-hot encoded vectors representing LLM judgments can be replaced with the judgment probabilities derived from logits values. This modification could enable potentially finer alignment between LLM outputs and human evaluations.
Furthermore, our alignment method can be extended to integrate other supervised learning techniques or objectives for handling more complex tasks, or to condition alignment on model inputs for improved performance where necessary.
We leave the exploration of these directions to future work, as they offer promising avenues for enhancing model alignment and performance, particularly in more complex or input-conditioned tasks.

Additionally, our approach lays the groundwork for multi-model alignment, enabling the learned mappings to be adapted for improving agreement across different LLMs. This adaptation could facilitate the transfer of knowledge between models, enhancing their reliability in judgment tasks. For instance, we could explore concatenating the outputs of multiple models and mapping them to human labels, effectively combining the strengths of various language models. Such a strategy would not only leverage diverse insights but also improve overall performance in alignment tasks. Related work, such as that proposed by \citet{verga2024replacing}, investigates multi-model alignment strategies that could complement and extend our findings, opening avenues for more robust evaluation frameworks.

Finally, our work allows LLMs to utilize different judgment spaces than human evaluators. However, we leave the exploration of this aspect to future work.
Enabling LLMs to have judgment spaces that differ from human evaluators offers several advantages. We can employ prompt optimization techniques to discover the optimal judgment space for an LLM. This allows us to align the model's outputs with human judgments, thereby achieving better agreement. In this way, we can search for the best judgment space for the model while preserving the human judgments as they are.

\section{Limitations}
\label{sec:limitation}
While our proposed alignment approach demonstrates significant improvements in aligning LLM judgments with human evaluations, there are some limitations to consider. 

First, the effectiveness of our method may vary depending on the specific characteristics of the domain and tasks involved. Some domains may present more complex alignment challenges and require more complicated alignment approaches, which may need larger number of training examples as a potential downside.

Second, while we compare our approach with in-context learning, we did not comprehensively explore the prompt space to see if we can improve the agreement and alignment through prompt optimisation. 
Instead, we used the prompts provided along with the task datasets from Judge-Bench \citep{bavaresco2024llms} to demonstrate the improvement in agreement through our approach. 
We believe prompt optimisation is orthogonal to our approach and both can be used at the same time. 
Furthermore, the flexibility of our approach for using different judgment spaces than human evaluators expands the exploration space for the prompt optimisation and may lead to better results.
Although we did not explicitly test these directions in our current work, they present an exciting opportunity for future research.

Despite these considerations, our findings provide a solid foundation for future research and practical applications, demonstrating the potential of our simple alignment method to enhance LLM as a judge performance across various domains.



\bibliography{references}

\begin{thebibliography}{42}
\providecommand{\natexlab}[1]{#1}

\bibitem[{Abercrombie and Rieser(2022)}]{abercrombie2022risk}
Gavin Abercrombie and Verena Rieser. 2022.
\newblock \href {https://aclanthology.org/2022.aacl-short.30} {Risk-graded
  safety for handling medical queries in conversational {AI}}.
\newblock In \emph{Proceedings of the 2nd Conference of the Asia-Pacific
  Chapter of the Association for Computational Linguistics and the 12th
  International Joint Conference on Natural Language Processing (Volume 2:
  Short Papers)}, pages 234--243, Online only. Association for Computational
  Linguistics.

\bibitem[{Anthropic(2024)}]{anthropic2024claude3}
Anthropic. 2024.
\newblock The {Claude} 3 model family: {Opus}, {Sonnet}, {Haiku}.
\newblock \url{https://anthropic.com/claude-3-model-card}.
\newblock Retrieved Oct~2024.

\bibitem[{Bavaresco et~al.(2024)Bavaresco, Bernardi, Bertolazzi, Elliott,
  Fernández, Gatt, Ghaleb, Giulianelli, Hanna, Koller, Martins, Mondorf,
  Neplenbroek, Pezzelle, Plank, Schlangen, Suglia, Surikuchi, Takmaz, and
  Testoni}]{bavaresco2024llms}
Anna Bavaresco, Raffaella Bernardi, Leonardo Bertolazzi, Desmond Elliott,
  Raquel Fernández, Albert Gatt, Esam Ghaleb, Mario Giulianelli, Michael
  Hanna, Alexander Koller, André F.~T. Martins, Philipp Mondorf, Vera
  Neplenbroek, Sandro Pezzelle, Barbara Plank, David Schlangen, Alessandro
  Suglia, Aditya~K Surikuchi, Ece Takmaz, and Alberto Testoni. 2024.
\newblock \href {https://arxiv.org/abs/2406.18403} {{LLMs} instead of human
  judges? {A} large scale empirical study across 20 {NLP} evaluation tasks}.
\newblock \emph{Preprint}, arXiv:2406.18403.

\bibitem[{Brown et~al.(2020)Brown, Mann, Ryder, Subbiah, Kaplan, Dhariwal,
  Neelakantan, Shyam, Sastry, Askell, Agarwal, Herbert-Voss, Krueger, Henighan,
  Child, Ramesh, Ziegler, Wu, Winter, Hesse, Chen, Sigler, teusz Litwin, Gray,
  Chess, Clark, Berner, McCandlish, Radford, Sutskever, and
  Amodei}]{brown2020language}
Tom~B. Brown, Benjamin Mann, Nick Ryder, Melanie Subbiah, Jared Kaplan,
  Prafulla Dhariwal, Arvind Neelakantan, Pranav Shyam, Girish Sastry, Amanda
  Askell, Sandhini Agarwal, Ariel Herbert-Voss, Gretchen Krueger, Tom Henighan,
  Rewon Child, Aditya Ramesh, Daniel~M. Ziegler, Jeff Wu, Clemens Winter,
  Christopher Hesse, Mark Chen, Eric Sigler, Ma~teusz Litwin, Scott Gray,
  Benjamin Chess, Jack Clark, Christopher Berner, Sam McCandlish, Alec Radford,
  Ilya Sutskever, and Dario Amodei. 2020.
\newblock \href {https://arxiv.org/abs/2005.14165} {Language models are
  few-shot learners}.
\newblock \emph{Preprint}, ArXiv:2005.14165.

\bibitem[{Camburu et~al.(2018)Camburu, Rockt\"{a}schel, Lukasiewicz, and
  Blunsom}]{camburu2018esnli}
Oana-Maria Camburu, Tim Rockt\"{a}schel, Thomas Lukasiewicz, and Phil Blunsom.
  2018.
\newblock \href
  {https://proceedings.neurips.cc/paper_files/paper/2018/file/4c7a167bb329bd92580a99ce422d6fa6-Paper.pdf}
  {{e-SNLI}: Natural language inference with natural language explanations}.
\newblock In \emph{Advances in Neural Information Processing Systems},
  volume~31. Curran Associates, Inc.

\bibitem[{Chiang and Lee(2023)}]{chiang2023large}
Cheng-Han Chiang and Hung-yi Lee. 2023.
\newblock \href {https://doi.org/10.18653/v1/2023.acl-long.870} {Can large
  language models be an alternative to human evaluations?}
\newblock In \emph{Proceedings of the 61st Annual Meeting of the Association
  for Computational Linguistics (Volume 1: Long Papers)}, pages 15607--15631,
  Toronto, Canada. Association for Computational Linguistics.

\bibitem[{Cobbe et~al.(2021)Cobbe, Kosaraju, Bavarian, Chen, Jun, Kaiser,
  Plappert, Tworek, Hilton, Nakano, Hesse, and Schulman}]{cobbe2021training}
Karl Cobbe, Vineet Kosaraju, Mohammad Bavarian, Mark Chen, Heewoo Jun, Lukasz
  Kaiser, Matthias Plappert, Jerry Tworek, Jacob Hilton, Reiichiro Nakano,
  Christopher Hesse, and John Schulman. 2021.
\newblock \href {https://arxiv.org/abs/2110.14168} {Training verifiers to solve
  math word problems}.
\newblock \emph{Preprint}, arXiv:2110.14168.

\bibitem[{Dua et~al.(2019)Dua, Wang, Dasigi, Stanovsky, Singh, and
  Gardner}]{dua2019drop}
Dheeru Dua, Yizhong Wang, Pradeep Dasigi, Gabriel Stanovsky, Sameer Singh, and
  Matt Gardner. 2019.
\newblock \href {https://doi.org/10.18653/v1/N19-1246} {{DROP}: A reading
  comprehension benchmark requiring discrete reasoning over paragraphs}.
\newblock In \emph{Proceedings of the 2019 Conference of the North {A}merican
  Chapter of the Association for Computational Linguistics: Human Language
  Technologies, Volume 1 (Long and Short Papers)}, pages 2368--2378,
  Minneapolis, Minnesota. Association for Computational Linguistics.

\bibitem[{Dubey et~al.(2024)Dubey, Jauhri, Pandey, Kadian, Al-Dahle, Letman,
  Mathur, Schelten, Yang, Fan et~al.}]{dubey2024llama}
Abhimanyu Dubey, Abhinav Jauhri, Abhinav Pandey, Abhishek Kadian, Ahmad
  Al-Dahle, Aiesha Letman, Akhil Mathur, Alan Schelten, Amy Yang, Angela Fan,
  et~al. 2024.
\newblock \href {https://arxiv.org/abs/2407.21783} {The {Llama} 3 herd of
  models}.
\newblock \emph{Preprint}, ArXiv:2407.21783.

\bibitem[{Faggioli et~al.(2023)Faggioli, Dietz, Clarke, Demartini, Hagen,
  Hauff, Kando, Kanoulas, Potthast, Stein, and
  Wachsmuth}]{faggioli2023perspectives}
Guglielmo Faggioli, Laura Dietz, Charles~L.A. Clarke, Gianluca Demartini,
  Matthias Hagen, Claudia Hauff, Noriko Kando, Evangelos Kanoulas, Martin
  Potthast, Benno Stein, and Henning Wachsmuth. 2023.
\newblock Perspectives on large language models for relevance judgment.
\newblock In \emph{Proceedings of the 2023 ACM SIGIR International Conference
  on Theory of Information Retrieval}, pages 39--50.

\bibitem[{Fei et~al.(2023)Fei, Hou, Chen, and
  Bosselut}]{fei-etal-2023-mitigating}
Yu~Fei, Yifan Hou, Zeming Chen, and Antoine Bosselut. 2023.
\newblock \href {https://doi.org/10.18653/v1/2023.acl-long.783} {Mitigating
  label biases for in-context learning}.
\newblock In \emph{Proceedings of the 61st Annual Meeting of the Association
  for Computational Linguistics (Volume 1: Long Papers)}, pages 14014--14031,
  Toronto, Canada. Association for Computational Linguistics.

\bibitem[{Freitag et~al.(2021)Freitag, Foster, Grangier, Ratnakar, Tan, and
  Macherey}]{freitag2021experts}
Markus Freitag, George Foster, David Grangier, Viresh Ratnakar, Qijun Tan, and
  Wolfgang Macherey. 2021.
\newblock \href {https://doi.org/10.1162/tacl_a_00437} {{Experts, Errors, and
  Context: A Large-Scale Study of Human Evaluation for Machine Translation}}.
\newblock \emph{Transactions of the Association for Computational Linguistics},
  9:1460--1474.

\bibitem[{Gilardi et~al.(2023)Gilardi, Alizadeh, and
  Kubli}]{gilardi2023chatgpt}
Fabrizio Gilardi, Meysam Alizadeh, and Ma{\"e}l Kubli. 2023.
\newblock {ChatGPT} outperforms crowd workers for text-annotation tasks.
\newblock \emph{Proceedings of the National Academy of Sciences},
  120(30):e2305016120.

\bibitem[{Golovneva et~al.(2023)Golovneva, Chen, Poff, Corredor, Zettlemoyer,
  Fazel-Zarandi, and Celikyilmaz}]{golovneva2023roscoe}
Olga Golovneva, Moya~Peng Chen, Spencer Poff, Martin Corredor, Luke
  Zettlemoyer, Maryam Fazel-Zarandi, and Asli Celikyilmaz. 2023.
\newblock \href {https://openreview.net/forum?id=xYlJRpzZtsY} {{ROSCOE}: A
  suite of metrics for scoring step-by-step reasoning}.
\newblock In \emph{The Eleventh International Conference on Learning
  Representations}.

\bibitem[{Grusky et~al.(2018)Grusky, Naaman, and Artzi}]{grusky2018newsroom}
Max Grusky, Mor Naaman, and Yoav Artzi. 2018.
\newblock \href {https://doi.org/10.18653/v1/N18-1065} {{N}ewsroom: A dataset
  of 1.3 million summaries with diverse extractive strategies}.
\newblock In \emph{Proceedings of the 2018 Conference of the North {A}merican
  Chapter of the Association for Computational Linguistics: Human Language
  Technologies, Volume 1 (Long Papers)}, pages 708--719, New Orleans,
  Louisiana. Association for Computational Linguistics.

\bibitem[{Hada et~al.(2024)Hada, Gumma, Wynter, Diddee, Ahmed, Choudhury, Bali,
  and Sitaram}]{hada2024large}
Rishav Hada, Varun Gumma, Adrian Wynter, Harshita Diddee, Mohamed Ahmed,
  Monojit Choudhury, Kalika Bali, and Sunayana Sitaram. 2024.
\newblock \href {https://aclanthology.org/2024.findings-eacl.71} {Are large
  language model-based evaluators the solution to scaling up multilingual
  evaluation?}
\newblock In \emph{Findings of the Association for Computational Linguistics:
  EACL 2024}, pages 1051--1070, St. Julian{'}s, Malta. Association for
  Computational Linguistics.

\bibitem[{Han et~al.(2023)Han, Hao, Dong, Sun, and Wei}]{han2023prototypical}
Zhixiong Han, Yaru Hao, Li~Dong, Yutao Sun, and Furu Wei. 2023.
\newblock \href {https://openreview.net/forum?id=nUsP9lFADUF} {Prototypical
  calibration for few-shot learning of language models}.
\newblock In \emph{The Eleventh International Conference on Learning
  Representations}.

\bibitem[{Hoerl and Kennard(1970)}]{hoerl1970ridge}
Arthur~E Hoerl and Robert~W Kennard. 1970.
\newblock Ridge regression: Biased estimation for nonorthogonal problems.
\newblock \emph{Technometrics}, 12(1):55--67.

\bibitem[{Huang et~al.(2024)Huang, Kwak, Park, and
  An}]{huang-etal-2024-chatgpt}
Fan Huang, Haewoon Kwak, Kunwoo Park, and Jisun An. 2024.
\newblock \href {https://aclanthology.org/2024.lrec-main.277} {{C}hat{GPT}
  rates natural language explanation quality like humans: But on which scales?}
\newblock In \emph{Proceedings of the 2024 Joint International Conference on
  Computational Linguistics, Language Resources and Evaluation (LREC-COLING
  2024)}, pages 3111--3132, Torino, Italia. ELRA and ICCL.

\bibitem[{Huang et~al.(2019)Huang, Le~Bras, Bhagavatula, and
  Choi}]{huang2019cosmos}
Lifu Huang, Ronan Le~Bras, Chandra Bhagavatula, and Yejin Choi. 2019.
\newblock \href {https://doi.org/10.18653/v1/D19-1243} {Cosmos {QA}: Machine
  reading comprehension with contextual commonsense reasoning}.
\newblock In \emph{Proceedings of the 2019 Conference on Empirical Methods in
  Natural Language Processing and the 9th International Joint Conference on
  Natural Language Processing (EMNLP-IJCNLP)}, pages 2391--2401, Hong Kong,
  China. Association for Computational Linguistics.

\bibitem[{Jackson and Messick(1958)}]{jackson1958content}
Douglas~N. Jackson and Samuel Messick. 1958.
\newblock Content and style in personality assessment.
\newblock \emph{Psychological Bulletin}, 55(4):243.

\bibitem[{Jiang et~al.(2024)Jiang, Sablayrolles, Roux, Mensch, Savary, Bamford,
  Chaplot, de~Las~Casas, Hanna, Bressand, Lengyel, Bour, Lample, Lavaud,
  Saulnier, Lachaux, Stock, Subramanian, Yang, Antoniak, Scao, Gervet, Lavril,
  Wang, Lacroix, and Sayed}]{jiang2024mixtral}
Albert~Q. Jiang, Alexandre Sablayrolles, Antoine Roux, Arthur Mensch, Blanche
  Savary, Chris Bamford, Devendra~Singh Chaplot, Diego de~Las~Casas, Emma~Bou
  Hanna, Florian Bressand, Gianna Lengyel, Guillaume Bour, Guillaume Lample,
  L'elio~Renard Lavaud, Lucile Saulnier, Marie-Anne Lachaux, Pierre Stock,
  Sandeep Subramanian, Sophia Yang, Szymon Antoniak, Teven~Le Scao,
  Th{\'e}ophile Gervet, Thibaut Lavril, Thomas Wang, Timoth{\'e}e Lacroix, and
  William~El Sayed. 2024.
\newblock \href {https://arxiv.org/abs/2401.04088} {Mixtral of experts}.
\newblock \emph{Preprint}, ArXiv:2401.04088.

\bibitem[{Kocmi and Federmann(2023)}]{kocmi2023large}
Tom Kocmi and Christian Federmann. 2023.
\newblock \href {https://aclanthology.org/2023.eamt-1.19} {Large language
  models are state-of-the-art evaluators of translation quality}.
\newblock In \emph{Proceedings of the 24th Annual Conference of the European
  Association for Machine Translation}, pages 193--203, Tampere, Finland.
  European Association for Machine Translation.

\bibitem[{Koo et~al.(2024)Koo, Lee, Raheja, Park, Kim, and
  Kang}]{koo2024benchmarking}
Ryan Koo, Minhwa Lee, Vipul Raheja, Jong~Inn Park, Zae~Myung Kim, and Dongyeop
  Kang. 2024.
\newblock \href {https://doi.org/10.18653/v1/2024.findings-acl.29}
  {Benchmarking cognitive biases in large language models as evaluators}.
\newblock In \emph{Findings of the Association for Computational Linguistics
  ACL 2024}, pages 517--545, Bangkok, Thailand and virtual meeting. Association
  for Computational Linguistics.

\bibitem[{Li et~al.(2022)Li, Sharma, Lu, Cheung, and Reddy}]{li2022using}
Zichao Li, Prakhar Sharma, Xing~Han Lu, Jackie~CK Cheung, and Siva Reddy. 2022.
\newblock \href {https://arxiv.org/abs/2204.03025} {Using interactive feedback
  to improve the accuracy and explainability of question answering systems
  post-deployment}.
\newblock \emph{Preprint}, arXiv:2204.03025.

\bibitem[{Liu et~al.(2023)Liu, Iter, Xu, Wang, Xu, and Zhu}]{liu2023geval}
Yang Liu, Dan Iter, Yichong Xu, Shuohang Wang, Ruochen Xu, and Chenguang Zhu.
  2023.
\newblock \href {https://doi.org/10.18653/v1/2023.emnlp-main.153} {{G}-eval:
  {NLG} evaluation using {GPT}-4 with better human alignment}.
\newblock In \emph{Proceedings of the 2023 Conference on Empirical Methods in
  Natural Language Processing}, pages 2511--2522, Singapore. Association for
  Computational Linguistics.

\bibitem[{Naismith et~al.(2023)Naismith, Mulcaire, and
  Burstein}]{naismith2023automated}
Ben Naismith, Phoebe Mulcaire, and Jill Burstein. 2023.
\newblock Automated evaluation of written discourse coherence using {GPT-4}.
\newblock In \emph{Proceedings of the 18th Workshop on Innovative Use of NLP
  for Building Educational Applications (BEA 2023)}, pages 394--403.

\bibitem[{Paulhus(1991)}]{paulhus1991measurement}
Delroy~L. Paulhus. 1991.
\newblock \href {https://doi.org/10.1016/B978-0-12-590241-0.50006-X}
  {Measurement and control of response bias}.
\newblock In John~P. Robinson, Phillip~R. Shaver, and Lawrence~S. Wrightsman,
  editors, \emph{Measures of Personality and Social Psychological Attitudes},
  pages 17--59. Academic Press.

\bibitem[{Pavlovic and Poesio(2024)}]{pavlovic2024effectiveness}
Maja Pavlovic and Massimo Poesio. 2024.
\newblock \href {https://aclanthology.org/2024.nlperspectives-1.11} {The
  effectiveness of {LLM}s as annotators: A comparative overview and empirical
  analysis of direct representation}.
\newblock In \emph{Proceedings of the 3rd Workshop on Perspectivist Approaches
  to NLP (NLPerspectives) @ LREC-COLING 2024}, pages 100--110, Torino, Italia.
  ELRA and ICCL.

\bibitem[{Reif and Schwartz(2024)}]{reif2024beyond}
Yuval Reif and Roy Schwartz. 2024.
\newblock \href {https://doi.org/10.18653/v1/2024.naacl-long.378} {Beyond
  performance: Quantifying and mitigating label bias in {LLM}s}.
\newblock In \emph{Proceedings of the 2024 Conference of the North American
  Chapter of the Association for Computational Linguistics: Human Language
  Technologies (Volume 1: Long Papers)}, pages 6784--6798, Mexico City, Mexico.
  Association for Computational Linguistics.

\bibitem[{Schoonees et~al.(2015)Schoonees, Van~de Velden, and
  Groenen}]{schoonees2015constrained}
Pieter~C. Schoonees, Michel Van~de Velden, and Patrick J.~F. Groenen. 2015.
\newblock \href {https://doi.org/10.1007/s11336-015-9458-9} {Constrained dual
  scaling for detecting response styles in categorical data}.
\newblock \emph{Psychometrika}, 80:968--994.

\bibitem[{Tjuatja et~al.(2024)Tjuatja, Chen, Wu, Talwalkwar, and
  Neubig}]{tjuatja2024llms}
Lindia Tjuatja, Valerie Chen, Tongshuang Wu, Ameet Talwalkwar, and Graham
  Neubig. 2024.
\newblock \href {https://doi.org/10.1162/tacl_a_00685} {{Do {LLMs} Exhibit
  Human-like Response Biases? {A} Case Study in Survey Design}}.
\newblock \emph{Transactions of the Association for Computational Linguistics},
  12:1011--1026.

\bibitem[{T{\"o}rnberg(2023)}]{tornberg2023chatgpt}
Petter T{\"o}rnberg. 2023.
\newblock \href {https://arxiv.org/abs/2304.06588} {{ChatGPT-4} outperforms
  experts and crowd workers in annotating political {Twitter} messages with
  zero-shot learning}.
\newblock \emph{Preprint}, arXiv:2304.06588.

\bibitem[{Van~Rosmalen et~al.(2010)Van~Rosmalen, Van~Herk, and
  Groenen}]{van2010identifying}
Joost Van~Rosmalen, Hester Van~Herk, and Patrick J.~F. Groenen. 2010.
\newblock \href {https://www.jstor.org/stable/20618962} {Identifying response
  styles: A latent-class bilinear multinomial logit model}.
\newblock \emph{Journal of Marketing Research}, 47(1):157--172.

\bibitem[{Verga et~al.(2024)Verga, Hofstatter, Althammer, Su, Piktus,
  Arkhangorodsky, Xu, White, and Lewis}]{verga2024replacing}
Pat Verga, Sebastian Hofstatter, Sophia Althammer, Yixuan Su, Aleksandra
  Piktus, Arkady Arkhangorodsky, Minjie Xu, Naomi White, and Patrick Lewis.
  2024.
\newblock \href {https://arxiv.org/abs/2404.18796} {Replacing judges with
  juries: Evaluating {LLM} generations with a panel of diverse models}.
\newblock \emph{Preprint}, arXiv:2404.18796.

\bibitem[{Wang et~al.(2023)Wang, Li, Chen, Cai, Zhu, Lin, Cao, Liu, Liu, and
  Sui}]{wang2023large}
Peiyi Wang, Lei Li, Liang Chen, Zefan Cai, Dawei Zhu, Binghuai Lin, Yunbo Cao,
  Qi~Liu, Tianyu Liu, and Zhifang Sui. 2023.
\newblock \href {https://arxiv.org/abs/2305.17926} {Large language models are
  not fair evaluators}.
\newblock \emph{Preprint}, arXiv:2305.17926.

\bibitem[{Wu and Aji(2023)}]{wu2023style}
Minghao Wu and Alham~Fikri Aji. 2023.
\newblock \href {https://arxiv.org/abs/2307.03025} {Style over substance:
  Evaluation biases for large language models}.
\newblock \emph{Preprint}, arXiv:2307.03025.

\bibitem[{Zeng et~al.(2024)Zeng, Yu, Gao, Meng, Goyal, and
  Chen}]{zeng2024evaluating}
Zhiyuan Zeng, Jiatong Yu, Tianyu Gao, Yu~Meng, Tanya Goyal, and Danqi Chen.
  2024.
\newblock \href {https://openreview.net/forum?id=tr0KidwPLc} {Evaluating large
  language models at evaluating instruction following}.
\newblock In \emph{The Twelfth International Conference on Learning
  Representations}.

\bibitem[{Zhao et~al.(2021)Zhao, Wallace, Feng, Klein, and
  Singh}]{zhao2021calibrate}
Zihao Zhao, Eric Wallace, Shi Feng, Dan Klein, and Sameer Singh. 2021.
\newblock \href {https://proceedings.mlr.press/v139/zhao21c.html} {Calibrate
  before use: Improving few-shot performance of language models}.
\newblock In \emph{Proceedings of the 38th International Conference on Machine
  Learning}, pages 12697--12706. PMLR.

\bibitem[{Zheng et~al.(2023{\natexlab{a}})Zheng, Zhou, Meng, Zhou, and
  Huang}]{zheng2023large}
Chujie Zheng, Hao Zhou, Fandong Meng, Jie Zhou, and Minlie Huang.
  2023{\natexlab{a}}.
\newblock Large language models are not robust multiple choice selectors.
\newblock In \emph{The Twelfth International Conference on Learning
  Representations}.

\bibitem[{Zheng et~al.(2023{\natexlab{b}})Zheng, Chiang, Sheng, Zhuang, Wu,
  Zhuang, Lin, Li, Li, Xing et~al.}]{zheng2023judging}
Lianmin Zheng, Wei-Lin Chiang, Ying Sheng, Siyuan Zhuang, Zhanghao Wu, Yonghao
  Zhuang, Zi~Lin, Zhuohan Li, Dacheng Li, Eric Xing, et~al. 2023{\natexlab{b}}.
\newblock Judging {LLM-as-a-Judge} with {MT}-{B}ench and {C}hatbot {A}rena.
\newblock \emph{Advances in Neural Information Processing Systems},
  36:46595--46623.

\bibitem[{Zhou et~al.(2024)Zhou, Wan, Proleev, Mincu, Chen, Heller, and
  Roy}]{zhou2024batch}
Han Zhou, Xingchen Wan, Lev Proleev, Diana Mincu, Jilin Chen, Katherine~A
  Heller, and Subhrajit Roy. 2024.
\newblock \href {https://openreview.net/forum?id=L3FHMoKZcS} {Batch
  calibration: Rethinking calibration for in-context learning and prompt
  engineering}.
\newblock In \emph{The Twelfth International Conference on Learning
  Representations}.

\end{thebibliography}

\newpage
\appendix
\section{Limited Training Examples}
In Table~\ref{tab:alignment_results_20_samples}, we demonstrate the performance of our approach when the number of training examples are limited to $20$ samples while $300$ samples are used for testing. We used $25$\% of the dataset as training examples if the total number of examples in the dataset less then $320$ samples.
The table indicates that our approach can also work in extremely low data regimes.

\begin{table*}[t]
\hspace{-0.5in}
\footnotesize

\begin{tabular}{@{}lcrrrrrr@{}}
\toprule
 &  & \multicolumn{3}{c}{\textbf{Non-aligned}} & \multicolumn{3}{c}{\textbf{Aligned}} \\
\cmidrule(lr){3-5} \cmidrule(lr){6-8}
 &  & {\scshape\scriptsize Claude-3} & {\scshape\scriptsize Mixtral 8x7B} & {\scshape\scriptsize Llama-3 70B} & {\scshape\scriptsize Claude-3} & {\scshape\scriptsize Mixtral 8x7B} & {\scshape\scriptsize Llama-3 70B} \\
\textbf{Task Name} & {{\scshape\scriptsize Human}} & {\scshape\scriptsize Sonnet} & {\scshape\scriptsize Instruct} & {\scshape\scriptsize Instruct} & {\scshape\scriptsize Sonnet} & {\scshape\scriptsize Instruct} & {\scshape\scriptsize Instruct} \\
\midrule
Feedback-QA & 44.50 & 42.11 {\tiny $\pm 2.16$} & 43.06 {\tiny $\pm 1.57$} & 47.38 {\tiny $\pm 2.10$} & \underline{48.88} {\tiny $\pm 3.47$} & \underline{48.31} {\tiny $\pm 3.97$} & \underline{\textbf{49.19}} {\tiny $\pm 4.26$} \\
LLMBar Natural & -- & \underline{85.07} {\tiny $\pm 1.67$} & \underline{77.20} {\tiny $\pm 2.19$} & \underline{\textbf{86.52}} {\tiny $\pm 1.21$} & \underline{85.07} {\tiny $\pm 1.67$} & \underline{77.20} {\tiny $\pm 2.19$} & \underline{\textbf{86.52}} {\tiny $\pm 1.21$} \\
Medical Safety \\
\quad \emph{Query Risk Level} & -- & 39.67 {\tiny $\pm 1.15$} & 43.20 {\tiny $\pm 1.30$} & 21.12 {\tiny $\pm 1.80$} & \underline{\textbf{84.03}} {\tiny $\pm 2.09$} & \underline{83.93} {\tiny $\pm 3.36$} & \underline{83.50} {\tiny $\pm 3.50$} \\
\quad \emph{Response Type} & -- & 5.63 {\tiny $\pm 0.52$} & 11.03 {\tiny $\pm 1.31$} & 5.63 {\tiny $\pm 1.00$} & \underline{\textbf{75.22}} {\tiny $\pm 6.03$} & \underline{64.76} {\tiny $\pm 4.51$} & \underline{72.95} {\tiny $\pm 5.32$} \\
Newsroom \\
\quad \emph{Coherence} & 24.29 & 27.87 {\tiny $\pm 0.96$} & \underline{26.11} {\tiny $\pm 0.50$} & \underline{\textbf{31.94}} {\tiny $\pm 0.58$} & \underline{29.22} {\tiny $\pm 3.63$} & 25.95 {\tiny $\pm 3.16$} & 28.74 {\tiny $\pm 3.10$} \\
\quad \emph{Fluency} & 21.35 & \underline{28.03} {\tiny $\pm 0.65$} & 20.27 {\tiny $\pm 0.82$} & \underline{\textbf{30.70}} {\tiny $\pm 0.70$} & 22.80 {\tiny $\pm 2.95$} & \underline{20.82} {\tiny $\pm 2.37$} & 25.42 {\tiny $\pm 2.60$} \\
\quad \emph{Informativeness} & 31.75 & 32.77 {\tiny $\pm 0.86$} & 15.53 {\tiny $\pm 1.21$} & \underline{\textbf{35.01}} {\tiny $\pm 0.91$} & \underline{32.90} {\tiny $\pm 2.81$} & \underline{24.32} {\tiny $\pm 3.21$} & 32.76 {\tiny $\pm 3.63$} \\
\quad \emph{Relevance} & 30.71 & \underline{30.96} {\tiny $\pm 0.93$} & 21.24 {\tiny $\pm 0.58$} & \underline{\textbf{33.58}} {\tiny $\pm 0.68$} & 29.90 {\tiny $\pm 1.56$} & \underline{26.85} {\tiny $\pm 5.01$} & 32.26 {\tiny $\pm 3.36$} \\
ROSCOE-Cosmos \\
\quad \emph{Coherency} & -- & 42.65 {\tiny $\pm 1.90$} & 20.41 {\tiny $\pm 1.61$} & 36.92 {\tiny $\pm 1.95$} & \underline{\textbf{49.12}} {\tiny $\pm 3.07$} & \underline{35.92} {\tiny $\pm 6.48$} & \underline{48.81} {\tiny $\pm 6.91$} \\
\quad \emph{Contradiction} & -- & 68.64 {\tiny $\pm 2.54$} & 57.20 {\tiny $\pm 1.22$} & 61.55 {\tiny $\pm 2.17$} & \underline{76.19} {\tiny $\pm 3.87$} & \underline{\textbf{77.94}} {\tiny $\pm 0.98$} & \underline{74.19} {\tiny $\pm 7.63$} \\
\quad \emph{Missing Steps} & -- & \underline{59.12} {\tiny $\pm 1.76$} & \underline{56.73} {\tiny $\pm 1.37$} & \underline{\textbf{61.72}} {\tiny $\pm 2.20$} & 56.80 {\tiny $\pm 2.57$} & 54.08 {\tiny $\pm 4.12$} & 59.40 {\tiny $\pm 2.40$} \\
\quad \emph{Overall Quality} & -- & 29.86 {\tiny $\pm 2.38$} & 18.50 {\tiny $\pm 1.48$} & 33.47 {\tiny $\pm 1.98$} & \underline{39.59} {\tiny $\pm 9.34$} & \underline{30.41} {\tiny $\pm 4.61$} & \underline{\textbf{41.62}} {\tiny $\pm 5.83$} \\
ROSCOE-DROP \\
\quad \emph{Coherency} & -- & 72.66 {\tiny $\pm 1.98$} & 34.68 {\tiny $\pm 2.40$} & 67.56 {\tiny $\pm 1.75$} & \underline{\textbf{75.57}} {\tiny $\pm 2.30$} & \underline{70.95} {\tiny $\pm 9.04$} & \underline{74.23} {\tiny $\pm 1.44$} \\
\quad \emph{Contradiction} & -- & 88.16 {\tiny $\pm 0.70$} & 71.24 {\tiny $\pm 1.39$} & 86.48 {\tiny $\pm 0.96$} & \underline{92.03} {\tiny $\pm 2.96$} & \underline{\textbf{94.03}} {\tiny $\pm 0.65$} & \underline{93.69} {\tiny $\pm 2.79$} \\
\quad \emph{Missing Steps} & -- & \underline{51.46} {\tiny $\pm 1.70$} & \underline{51.58} {\tiny $\pm 1.34$} & \underline{\textbf{55.44}} {\tiny $\pm 1.59$} & 50.76 {\tiny $\pm 2.23$} & 48.55 {\tiny $\pm 2.18$} & 48.96 {\tiny $\pm 4.60$} \\
\quad \emph{Overall Quality} & -- & \underline{43.42} {\tiny $\pm 2.14$} & \underline{40.89} {\tiny $\pm 1.82$} & \underline{\textbf{43.53}} {\tiny $\pm 2.20$} & 42.22 {\tiny $\pm 10.60$} & 36.20 {\tiny $\pm 8.57$} & 42.68 {\tiny $\pm 8.48$} \\
ROSCOE-ESNLI \\
\quad \emph{Coherency} & -- & 84.65 {\tiny $\pm 1.19$} & 39.39 {\tiny $\pm 2.06$} & 73.59 {\tiny $\pm 1.80$} & \underline{85.96} {\tiny $\pm 2.32$} & \underline{\textbf{88.77}} {\tiny $\pm 0.94$} & \underline{85.77} {\tiny $\pm 6.95$} \\
\quad \emph{Contradiction} & -- & 93.25 {\tiny $\pm 0.96$} & 50.57 {\tiny $\pm 2.37$} & 93.46 {\tiny $\pm 0.99$} & \underline{95.09} {\tiny $\pm 1.63$} & \underline{\textbf{96.55}} {\tiny $\pm 0.84$} & \underline{95.05} {\tiny $\pm 2.02$} \\
\quad \emph{Missing Steps} & -- & 67.98 {\tiny $\pm 1.97$} & 27.28 {\tiny $\pm 1.38$} & 64.26 {\tiny $\pm 1.47$} & \underline{72.11} {\tiny $\pm 2.44$} & \underline{\textbf{72.98}} {\tiny $\pm 1.46$} & \underline{70.68} {\tiny $\pm 4.32$} \\
\quad \emph{Overall Quality} & -- & \underline{\textbf{48.07}} {\tiny $\pm 2.63$} & 38.86 {\tiny $\pm 2.26$} & \underline{47.55} {\tiny $\pm 2.73$} & 46.84 {\tiny $\pm 2.99$} & \underline{39.21} {\tiny $\pm 7.92$} & 43.63 {\tiny $\pm 5.30$} \\
ROSCOE-GSM8K \\
\quad \emph{Coherency} & -- & 59.73 {\tiny $\pm 1.84$} & 47.50 {\tiny $\pm 2.70$} & \underline{\textbf{63.89}} {\tiny $\pm 1.83$} & \underline{62.60} {\tiny $\pm 2.01$} & \underline{56.53} {\tiny $\pm 5.42$} & 63.76 {\tiny $\pm 2.30$} \\
\quad \emph{Contradiction} & -- & 80.67 {\tiny $\pm 0.94$} & 55.30 {\tiny $\pm 2.37$} & 80.53 {\tiny $\pm 1.42$} & \underline{82.13} {\tiny $\pm 1.60$} & \underline{\textbf{83.32}} {\tiny $\pm 1.61$} & \underline{83.26} {\tiny $\pm 2.32$} \\
\quad \emph{Missing Steps} & -- & \underline{81.13} {\tiny $\pm 1.46$} & 42.77 {\tiny $\pm 0.77$} & \underline{\textbf{84.09}} {\tiny $\pm 1.26$} & \underline{81.13} {\tiny $\pm 1.46$} & \underline{52.66} {\tiny $\pm 10.38$} & \underline{\textbf{84.09}} {\tiny $\pm 1.26$} \\
\quad \emph{Overall Quality} & -- & \underline{\textbf{71.60}} {\tiny $\pm 2.15$} & 49.59 {\tiny $\pm 2.37$} & 65.46 {\tiny $\pm 2.03$} & 71.33 {\tiny $\pm 4.00$} & \underline{57.02} {\tiny $\pm 5.07$} & \underline{70.89} {\tiny $\pm 2.52$} \\
SummEval \\
\quad \emph{Coherence} & -- & 31.47 {\tiny $\pm 1.71$} & \underline{28.79} {\tiny $\pm 2.06$} & \underline{\textbf{40.90}} {\tiny $\pm 1.48$} & \underline{37.13} {\tiny $\pm 4.46$} & 28.52 {\tiny $\pm 3.91$} & 35.65 {\tiny $\pm 4.42$} \\
\quad \emph{Consistency} & -- & \underline{\textbf{83.64}} {\tiny $\pm 1.53$} & 68.84 {\tiny $\pm 1.80$} & 68.82 {\tiny $\pm 2.53$} & 82.54 {\tiny $\pm 1.72$} & \underline{81.17} {\tiny $\pm 3.28$} & \underline{71.24} {\tiny $\pm 5.99$} \\
\quad \emph{Fluency} & -- & 4.50 {\tiny $\pm 0.83$} & 3.33 {\tiny $\pm 0.58$} & 2.58 {\tiny $\pm 0.33$} & \underline{77.17} {\tiny $\pm 6.18$} & \underline{\textbf{77.53}} {\tiny $\pm 3.84$} & \underline{77.48} {\tiny $\pm 6.28$} \\
\quad \emph{Relevance} & -- & 37.27 {\tiny $\pm 1.53$} & 31.27 {\tiny $\pm 1.72$} & 41.34 {\tiny $\pm 1.40$} & \underline{\textbf{49.13}} {\tiny $\pm 8.17$} & \underline{44.33} {\tiny $\pm 8.00$} & \underline{41.37} {\tiny $\pm 11.08$} \\
WMT-20-EnDe & 30.02 & 19.48 {\tiny $\pm 1.45$} & 17.58 {\tiny $\pm 1.58$} & 17.15 {\tiny $\pm 1.45$} & \underline{\textbf{25.11}} {\tiny $\pm 3.03$} & \underline{23.29} {\tiny $\pm 2.25$} & \underline{20.40} {\tiny $\pm 3.10$} \\
\bottomrule
\end{tabular}

\caption{Repetition of the experiments in Table~\ref{tab:alignment_results} with only $20$ training examples (or $25$\% of the dataset if it is smaller than $320$ examples) instead of $100$ training examples. \label{tab:alignment_results_20_samples} }
\end{table*}

\section{Prompt Example}
Figure~\ref{fig:medical-safety-prompt-example} shows an example of the prompt we used in both our zero-shot and few-shot experiments for the \emph{medical safety (response type)} task.

\onecolumn
\begin{figure}[htbp]
\centering
\begin{lstlisting}
Read the conversation, which consists of:
    (1) a user query (usually to a medic), and
    (2) a chatbot response.

For the chatbot response, don't worry about whether or not the response is good medical advice.
We are only interested in what TYPE of response it is.
Select one of the following types:

- Non-medical query
- Irrelevant or nonsensical
- No answer
- General information
- Recommendations
- Treatment or diagnosis

{{ examples }}

# Example

## Query 
{{ query }}

## Response
{{ response }}

Response label:
    \end{lstlisting}

    \begin{lstlisting}
# Example {{ icl_example_i }}:

## Query 
{{ query }}

## Response
{{ response }}

Response label: {{ icl_example_label }}
\end{lstlisting}
\caption{Example (top) prompt for one of the judgment tasks (medical-safety: response type), where \texttt{\{\{ examples \}\}} is included for in-context learning experiments only. The \texttt{\{\{ examples \}\}} placeholder is replaced with one example for each output label following the example format (bottom) of the corresponding task.}
\label{fig:medical-safety-prompt-example}
\end{figure}
\twocolumn

\end{document}